\documentclass[review,12pt,authoryear]{elsarticle}
\usepackage[T1]{fontenc}
\usepackage{lmodern}
\usepackage{amsmath,amssymb,booktabs,graphicx,hyperref}
\biboptions{round,semicolon}
\journal{Engineering Applications of Artificial Intelligence}
\hypersetup{hidelinks,pdftitle={Reducing Multi-Horizon Photovoltaic Forecast Error through Tree-Anchored Expert Combination}}
\begin{document}
\begin{frontmatter}
\title{Reducing Multi-Horizon Photovoltaic Forecast Error through Tree-Anchored Expert Combination}
\author[1]{Sun Ze}
\author[1,2]{Zhou Liguo\corref{cor1}}
\author[1]{Xu Yuqing}
\author[1]{Yu Lei}
\author[1]{Jiang Mingming}
\author[2]{Alois Knoll}
\affiliation[1]{organization={School of Computer Science and Technology, Huaibei Normal University},country={China}}
\affiliation[2]{organization={Chair of Robotics, Artificial Intelligence and Real-time Systems, Technical University of Munich},country={Germany}}
\cortext[cor1]{Corresponding author. E-mail: zhouliguo@chnu.edu.cn}

\begin{abstract}
A strong direct tree model may remain the best standalone photovoltaic forecaster, yet a weighted combination can reduce its residual errors. We examine whether complementary temporal predictors can lower the tree's mean absolute error (MAE) over 15--240 minutes without weaker ensemble members degrading performance. The candidate bank contains LightGBM and seven neural implementations with different temporal representations. A combiner fitted to MAE has nonnegative weights that sum to one. Unhelpful candidates can receive zero weight, while the tree anchors the combination and provides a development-stage fallback. The main design shares weights across horizons; horizon-specific weights and alternative losses test whether further complexity is needed. On 56,078 windows from two co-located public PVDAQ systems, the historical shared-MAE policy reduces scaled MAE by 3.21\% relative to the tree.

The latest bounded neural correction does not reduce MAE further than this static combination. The results support selective combination of complementary predictions in this bank rather than equal averaging or indiscriminate addition of modules. Both analyses reuse a previously inspected evaluation period. They do not establish independent replication, weather-specific causation or universal gains from eight experts.
\end{abstract}
\end{frontmatter}
\section{Introduction}
We examine whether an MAE-aligned combination can reduce a strong direct tree's residual error across 16 forecast horizons. We predict photovoltaic AC power
from 15 to 240 minutes ahead. LightGBM is the reference because a method
that adds several neural networks should first demonstrate value over a
competitive tree under the same information and target definitions.
The goal is lower forecast error across this interval, not a larger model
catalogue or a better result on an unrelated task.

Power variability and the usefulness of recent observations can change
with forecast lead. Weather and verification choices also affect reported
solar-forecast performance \citep{antonanzas2016,yang2020}. These facts
motivate testing several temporal representations, but do not prove that
the errors in this dataset are caused by cloud motion. The present data
do not support assigning a separate cloud regime to each expert. Instead,
we test a narrower, measurable hypothesis: some forecasts that are worse
than the tree individually can still contribute useful residual
information when combined with it.

Related photovoltaic studies examine both temporal representations and
how their predictions are combined. \citet{costa2022convlstm} compares
convolutional, LSTM, and convolutional--LSTM networks at different
forecast horizons and studies transfer from geographically aggregated
training data to individual household systems.
\citet{zheng2025aws} combines kernel-clustering-based learner selection,
joint feature and hyperparameter optimization, and adaptive weighted
stacking for small-sample photovoltaic forecasting.
\citet{yang2025clustering} couples numerical-weather-prediction-based
time-series clustering with lag-enhanced gradient-boosting ensembles
for day-ahead forecasting. These studies motivate testing temporal
representations and fitted aggregation, but their datasets, information
sets, and horizons differ from those used here. We do not compare their
published error values directly with ours. The present study isolates
a narrower question: whether MAE-fitted static weights can extract
useful corrections to a strong direct tree, without adding a weather
clustering or joint feature-search module.

Simply averaging additional models can increase MAE if their own errors
outweigh any cancellation. Research on forecast combination and stacking supports estimating weights
from held-out predictions
\citep{bates1969,clemen1989,wang2023,wolpert1992,breiman1996}.
We use a bank of eight candidate models with a constrained linear combiner. The tree captures nonlinear relationships in engineered inputs, while the neural candidates represent temporal dependence in different ways. Nonnegative weights that sum to one allow a weak candidate to receive zero weight. Fitting those weights to absolute error aligns the combiner
with the paper's MAE target; squared-error fitting targets a different
functional \citep{gneiting2011}. No candidate is required to receive a positive weight.

The central question is: \emph{can an MAE-aligned, tree-anchored combination reduce a strong direct tree's residual error across 16 forecast horizons?} We first compare the combination against the tree, equal averaging and single-expert selection. Comparisons of fitting loss and weight granularity then examine the individual design choices. Error accounting checks
whether the observed benefit is compatible with useful member selection
and cancellation \citep{brown2005}; it is an explanation of the measured
errors, not a separate prediction task or a causal weather analysis.
The latest public retraining check tests the same accuracy question
without future observed weather and with matched training-label access.

This empirical study examines an existing candidate bank and a
restricted combiner. We do not claim a new stacking algorithm, an optimal number of experts, or an independently validated improvement over all single-model methods. Both historical and retraining analyses use an
evaluation period inspected in earlier research, and their protocols and
aggregation rules are reported separately.

\section{Prediction Task and Expert Roles}
\subsection{Eight Candidates for One Forecasting Task}
The primary bank uses PVDAQ systems 34 and 35, two co-located Nevada
systems, not two independent climate domains \citep{deline2021pvdaq}. Each
origin supplies 96 quarter-hour observations and produces 16 direct
predictions, covering 15--240 minutes. The tree uses a direct head for each
horizon. This is not a recursive experiment that repeatedly feeds the
tree's own forecasts back as observations.

\begin{figure}[tbp]
\centering
\includegraphics[width=\linewidth]{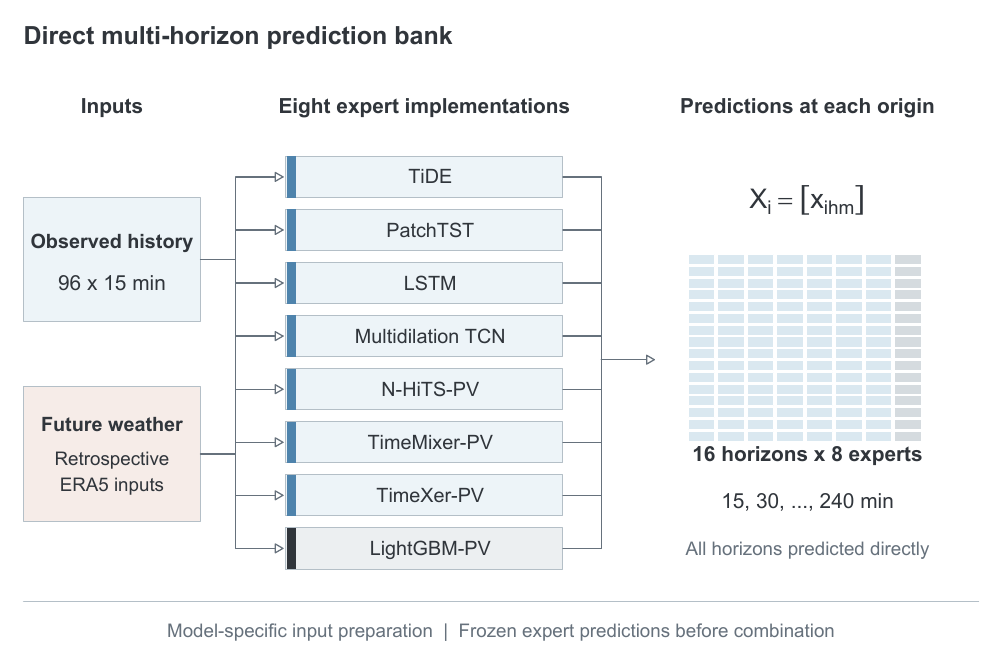}
\caption{Candidate representations for the same power-forecasting task.
Model-specific input preparation feeds seven neural implementations and
one direct tree predictor. The tree is the baseline and combination anchor;
the remaining forecasts are retained only through their fitted weights. Each produces
16 direct forecasts, giving a $16\times8$ prediction matrix at each origin.
The matrix cells show dimensions rather than measured values. Future weather comes from retrospective ERA5. All expert predictions are frozen before fitting
the combiner.}
\label{fig:architecture}
\end{figure}

Figure~\ref{fig:architecture} summarizes the prediction bank. It contains
project implementations based on TiDE \citep{das2023tide}, PatchTST
\citep{nie2023patchtst}, LSTM \citep{hochreiter1997}, temporal convolution
\citep{bai2018tcn}, N-HiTS \citep{challu2023nhits}, TimeMixer
\citep{wang2024timemixer}, and TimeXer \citep{wang2024timexer}, together
with a gradient-boosted tree predictor \citep{friedman2001,ke2017lightgbm}.
The project names include multidilation TCN, N-HiTS-PV, TimeMixer-PV,
TimeXer-PV, and LightGBM-PV. These labels identify local implementations;
we have not established reproduction of every published architecture.
All members predict the same power targets from the same origins; the
bank does not contain eight different forecasting tasks. Three historical neural seeds are 20260914,
20260915, and 20260916; the tree predictions are shared across these
evaluations. The neural models are trained for up to 20 epochs, batch size 256,
learning rate $10^{-4}$, weight decay $10^{-4}$, and MAE training, with
checkpoint choice on Epochval. The tree uses an L1 objective, at most 600 boosting rounds, a learning rate of 0.04, 31 leaves, a minimum child count of 64, an L2 coefficient of 1, and early stopping after 40 rounds. Before combination, member predictions are clipped to $[0,1.25]$. The tree branch has a defined role: sixteen direct LightGBM heads provide a strong tabular baseline using the available history and contextual features. Tree predictions are never fed back as observations. They also anchor the combination and provide a fallback if a candidate policy fails the subsequent development check. The tree is not included simply to bring the ensemble to eight members.

The neural candidates provide alternative temporal representations. TiDE uses a dense encoder--decoder, LSTM captures ordered recurrent dependence, and the multidilation TCN provides several local receptive fields. PatchTST groups observations into temporal patches, while the N-HiTS-PV and TimeMixer-PV implementations use different multiscale constructions. TimeXer-PV supplies a further contextual
attention representation. These are architectural motivations, not
demonstrated assignments to particular weather classes. The experiment
tests their combined value; it does not identify the causal contribution
of each internal component.

An expert here means a candidate predictor, not a weather-labelled
specialist or an input-dependent routing branch. Within a fitted policy,
weights are fixed across forecast origins. Several candidates may receive
zero weight. Thus the hypothesis is selective use of complementary
predictions, not that eight simultaneously active branches are necessary.

Power is divided by each system's observed training maximum. We call the
resulting metrics scaled MAE (sMAE) and scaled RMSE (sRMSE), not capacity
normalized errors or accuracy percentages. Targets are sampled AC power,
not interval-mean energy. We retain the scaling definition because error
measures are not interchangeable across normalization conventions
\citep{hyndman2006}. Finite negative targets are retained. Future
weather inputs are retrospective ERA5 values without operational forecast
issue times \citep{hersbach2020era5}. Consequently, this experiment does not
measure performance with genuinely issue-time-available numerical weather
predictions. The persistence baselines do not have this future-weather
information, which limits physical interpretation of that comparison.

For system $s$, let $b_s$ be the maximum finite training power in kW
($134.0$ and $116.6$ for systems 34 and 35). Observations and predictions
are expressed in units of $b_s$. With $H=16$, $N$ evaluation windows,
and $S=3$ seeds, the reported metrics are
\begin{align}
\mathrm{sMAE}&=\frac1S\sum_{r=1}^{S}\frac1{NH}
  \sum_{i,h}|q^{(r)}_{ih}-y_{ih}|,\\
\mathrm{sRMSE}&=\frac1S\sum_{r=1}^{S}
  \sqrt{\frac1{NH}\sum_{i,h}(q^{(r)}_{ih}-y_{ih})^2}.
\end{align}
These dimensionless metrics pool windows across systems before averaging
seed-level scores; sRMSE is not the RMSE of seed-averaged predictions.
Relative MAE reduction is $100(L_{\rm ref}-L_{\rm candidate})/L_{\rm ref}$,
not an accuracy-percentage-point increase. Horizon scores use the same
order of aggregation with $h$ fixed.

Every retained window has 96 historical and 16 future finite original
power samples; missing power is not interpolated. The frozen preparation rules exclude off-grid, ambiguous daylight-saving and duplicate timestamps. Missing standardized weather values are encoded as zero with a false validity mask; physical weather values remain NaN. Clipping member predictions to $[0,1.25]$ while retaining negative labels can affect errors near zero. All four combination conditions use the same retained windows and clipped prediction bank. These choices were not tuned in this comparison.

\subsection{Separation of Development Stages}
The prepared data include 20,053 training windows, 8,198 Epochval windows, 4,311 Fusionfit windows, 3,411 Selection windows, 5,756 Confirmation windows and 56,078 evaluation windows. Training uses data from 2014--2015; all subsequent development stages precede the 2016 evaluation period. Evaluation comprises 897,248 origin--horizon target entries. Overlapping windows, repeated targets, co-location and shared tree predictions prevent treating these entries or random seeds as independent replications.

Stages follow local calendar boundaries: training ends on 1 July 2015,
Epochval on 1 September, Fusionfit on 1 October, Selection on 1 November,
and Confirmation on 1 January 2016; evaluation ends on 1 January 2017.
Intervals are start-inclusive and end-exclusive. All 16 target timestamps
must belong to their assigned stage, although historical inputs may extend
into the preceding stage. Neither evaluation targets nor the evaluation
period's maximum power define the scaling constants.

Fusionfit estimates combination coefficients, Selection chooses the
policy, and Confirmation only accepts that choice or falls back to the
tree. Evaluation labels have no role in changing these choices. However, the
2016 period had already been inspected during earlier research. Freezing this execution does not remove the earlier research history. In particular, the integrated
baseline and optional-prefix controls were specified after the earlier
factorial evaluation. They are exploratory follow-up analyses, not a
preregistered confirmation \citep{nosek2018prereg}.

\section{Tree-Anchored MAE Reduction}
\subsection{Shared Weights as the Main Design}
\begin{figure}[tbp]
\centering
\includegraphics[width=\linewidth]{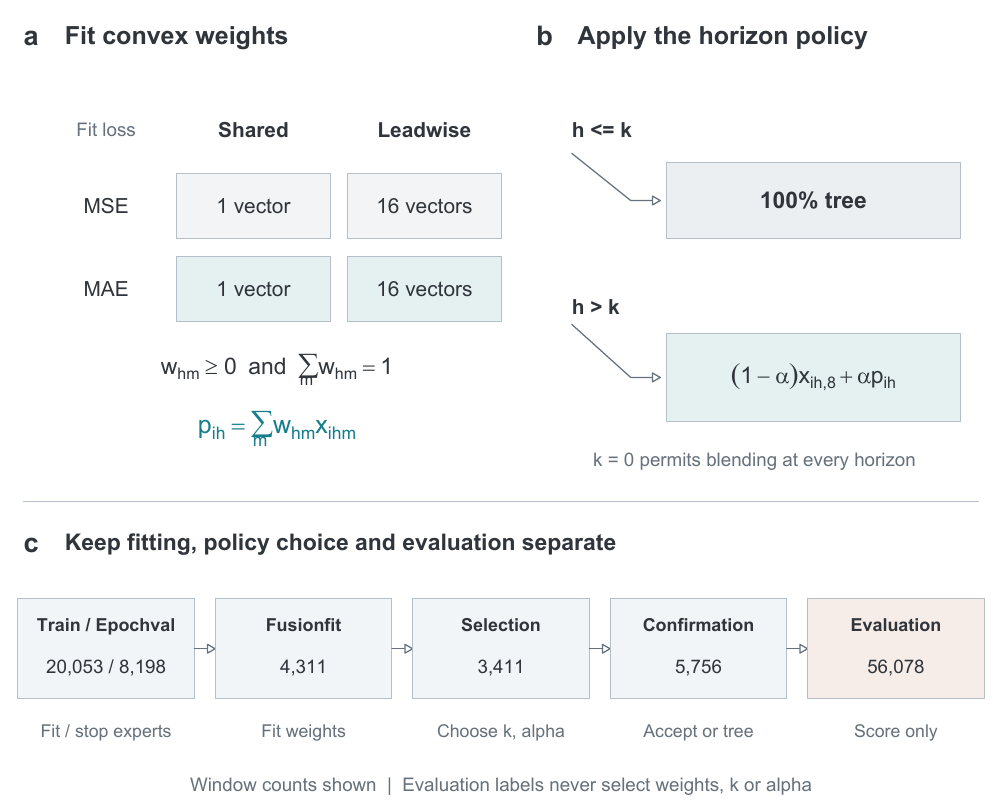}
\caption{Fusion and policy selection. (a) The factorial design crosses
fitting loss with shared or horizon-specific simplex weights.
(b) A policy uses the tree through horizon $k$ and blends it with the
fitted combination afterwards; $k=0$ removes the compulsory tree prefix.
(c) Distinct development stages fit the experts and weights, select
$(k,\alpha)$, and confirm the selected policy or return to the tree.
Confirmation does not search for a replacement candidate. Counts are
windows, not independent observations. Evaluation labels have no role in
these decisions, although the evaluation period was inspected in earlier
research.}
\label{fig:fusion}
\end{figure}

The shared-MAE policy is the main method evaluated in this paper. A
single simplex vector pools evidence across leads and limits the number
of fitted coefficients. Horizon-specific weights and MSE fitting are
controls for this design choice, not additional proposed methods.
For member predictions $x_{ihm}$, target $y_{ih}$, and weights
$w_h\in\mathcal S_8=\{w\geq0:\mathbf1^\top w=1\}$, define
\begin{equation}
 p_{ih}=\sum_{m=1}^8 w_{hm}x_{ihm}.
\end{equation}
The shared parameterization uses $w_h=w$ across all 16 horizons, giving seven free coefficients; the horizon-specific parameterization has 112. Each minimizes
either pooled squared error or pooled absolute error on Fusionfit, without
a ridge term. Separate leadwise fits minimize the corresponding
horizon-specific loss. MAE is solved by a linear program with feasibility
and objective checks. MSE uses SLSQP with an analytical gradient and an
unscaled simplex first-order optimality-gap tolerance of $10^{-8}$; the
numerical objective is multiplied by $10^4$ for solver conditioning.
Hereafter, \emph{horizon-specific} and \emph{leadwise} denote the same
parameterization. Explicitly, the shared objectives are
\begin{equation}
 \widehat w_{\ell}=\arg\min_{w\in\mathcal S_8}
 \frac1{N_F H}\sum_{i\in F,h}\ell\!\left(\sum_m w_m x_{ihm}-y_{ih}\right),
 \quad \ell(u)\in\{|u|,u^2\},
\end{equation}
where $F$ is Fusionfit and $N_F=4311$. Horizon-specific fitting replaces
the pooled objective by one objective per $h$. The MAE solver is HiGHS
dual simplex with default internal stopping settings; the post-solve
objective agreement tolerance is $10^{-7}$ and the residual-constraint
tolerance is $10^{-6}$. SLSQP uses \texttt{ftol}$=10^{-12}$ and at most
2000 iterations, followed by the optimality-gap check above. Solver failure
or a failed numerical check aborts the fitting run; it is not a tree
fallback. The latter is reserved for a completed policy that fails
Confirmation. No failed solve is silently replaced by a local fit.
Fit counts and execution provenance are provided in the supplement.
Table~\ref{tab:design} lists the four fitting conditions.
Figure~\ref{fig:fusion} separates coefficient fitting from the subsequent
policy decision.
\begin{table}[tbp]
\centering
\caption{Matched coefficient-fitting design. Every condition uses the same
eight-expert predictions and Fusionfit windows. These degrees of freedom count only simplex coefficients before boundary solutions. Policy selection is a separate step.}
\label{tab:design}
\small
\begin{tabular}{llrrl}
\toprule
Weights & Fit loss & Vectors & Free coeff. & Solver \\
\midrule
Shared & MSE & 1 & 7 & SLSQP \\
Shared & MAE & 1 & 7 & Linear program \\
Leadwise & MSE & 16 & 112 & SLSQP \\
Leadwise & MAE & 16 & 112 & Linear program \\
\bottomrule
\end{tabular}
\end{table}

Given fitted weights $w_h$, a policy with prefix $k$ and mixing strength
$\alpha$ predicts
\begin{equation}
 q_{ih}(k,\alpha)=
 \begin{cases}
 x_{ih,8}, & h\leq k,\\
 (1-\alpha)x_{ih,8}+\alpha p_{ih}, & h>k.
 \end{cases}
\end{equation}
The original policy grid contains pure tree plus
$k\in\{1,\ldots,15\}$ crossed with
$\alpha\in\{0.1,0.25,0.5,0.75,1\}$: 76 candidates per family and seed.
Selection minimizes sMAE; ties favor a longer tree prefix and a smaller mixing strength. Confirmation requires strictly lower aggregate
sMAE than the tree; it cannot choose a second-best candidate. We distinguish
\emph{full} fitted weights at all horizons, \emph{raw} weights with only
the first horizon forced to the tree, and the \emph{accepted} policy.
These versions must not be conflated when attributing an improvement to
the fitting objective.

To test the prefix restriction itself, the integration adds five $k=0$
candidates to each MAE family. The 81-candidate grid is selected and
confirmed remotely using the same stages. Selection also chooses a single
expert, either shared across horizons or separately by horizon;
Confirmation vetoes any non-improving choice at the same aggregation level.
Thus single-expert baselines are not chosen by their evaluation ranking.

The current shared-MAE implementation uses the pure tree for the first forecast and one shared vector for subsequent forecasts. This is an implementation constraint, not evidence that the switching time is universally optimal. The historical optional-prefix control permits $k=0$ and thus tests this restriction. Neither version estimates an origin-dependent weather router.

\subsection{Checking Why Combination Helps}
Taking the tree as the reference, let $e_{ihm}=x_{ihm}-y_{ih}$. Averaging over the evaluated origin--horizon entries, define
\begin{align}
 A(w)&=\mathbb E_{i,h}\left[\sum_m w_{hm}|e_{ihm}|-|e_{ih,8}|\right],\\
 C(w)&=\mathbb E_{i,h}\left[\sum_m w_{hm}|e_{ihm}|-
                         \left|\sum_m w_{hm}e_{ihm}\right|\right].
\end{align}
The triangle inequality gives $C\geq0$; direct subtraction then yields
\begin{equation}
 L(q)-L(\mathrm{tree})=A-C,
 \label{eq:ac}
\end{equation}
where $w$ represents the actual policy weights, including the tree mixture where applicable. This diagnostic describes realized loss, not causal effects. Both terms depend on labels, and cancellation on future targets is unknown at issue time. Development estimates can support a hypothesis but cannot ensure future acceptance. Maximizing cancellation alone may favor a prediction bank with larger member errors.

\subsection{Baselines and Uncertainty}
Besides the eight members and equal weights, we retain the original
ensemble as a historical baseline. It fits shared MSE weights with a
penalty toward uniform weights and uses a separate uniform-mixture
selection and confirmation rule. It is not the unregularized shared-MSE
factorial cell; its comparison does not isolate the fitting-loss effect.
The integration also evaluates raw and $[0,1.25]$-clipped persistence.
A specified smart-persistence proxy uses pvlib 0.13.1 Haurwitz clear-sky
global horizontal irradiance (GHI), recorded coordinates, and altitude
zero \citep{tato2019smart,pvlib2023}. It scales the last nonnegative power
by target/origin clear-sky GHI when origin GHI exceeds 5~W/m$^2$; otherwise
it retains that power. A zero target GHI gives zero output, and the final prediction is clipped to $[0,1.25]$. None of the thresholds is tuned using these evaluation results. This proxy lacks panel geometry and cloud advection
and is not a definitive physical baseline.

We average losses over seeds, not forecasts. Paired gains are calculated as reference sMAE minus candidate sMAE. Circular blocks of 7, 14 or 28 consecutive UTC calendar days resample the same origins jointly across methods, horizons and systems in 10,000 draws. Day counts weight the sampled losses. This construction retains local temporal dependence instead of resampling overlapping windows independently \citep{shao1993}.
Seven and 14 days are the existing analysis settings; 28 days provides a sensitivity check for longer dependence. All exceed the four-hour target
span, but that fact does not establish an optimal block length or remove
seasonal dependence. We do not select a block length by significance.
Reported percentile intervals are conditional on the fitted bank and decisions. They do not account for retraining variability, research-history selection or generalization to independent sites. Main-table and horizon-curve intervals
are pointwise and unadjusted. A separate retrospective sensitivity family
contains the four raw factorial contrasts (two loss contrasts and two
weight-granularity contrasts). Bonferroni percentile endpoints at
$0.05/(2\times4)$ and $1-0.05/(2\times4)$ give nominal 95\% family coverage
only to the extent that the underlying block intervals are valid. This
does not correct for earlier research decisions or simultaneous screening
of all horizons. We report effect sizes, do not label this analysis
confirmatory, and do not infer equivalence from an interval spanning zero.

\section{Results}
\subsection{Does Combination Beat the Strong Tree?}
Table~\ref{tab:baselines} compares policies on identical evaluation windows.
The accepted shared-MAE and horizon-specific-MAE policies yield sMAE values of 0.025420 and 0.025353, compared with 0.026262 for LightGBM. The relative reductions are 3.21\% and 3.46\%. Both shared and per-horizon singleton selectors retain the
tree at every horizon. Equal weights perform worse. The original ensemble
has the lowest sRMSE, 0.064109, compared with 0.064348 and 0.064398 for
the two accepted MAE policies. Thus the MAE gains do not establish
dominance on both metrics. Persistence controls are reported separately
in the supplement because they lack the bank's retrospective future weather.
\begin{table}[htbp]
\centering\small
\caption{Matched eight-member comparison. Entries average three seed-specific losses, not predictions. Positive gain denotes lower sMAE than the tree. Persistence controls are reported in the supplement.}\label{tab:baselines}
\begin{tabular}{lrrr}\toprule
Method & sMAE & sRMSE & Gain (\%) \\\midrule
Equal eight-member weights & 0.028874 & 0.066017 & -9.94 \\
Tree / accepted singletons & 0.026262 & 0.066494 & +0.00 \\
Original eight-member ensemble & 0.026095 & 0.064109 & +0.64 \\
Shared MSE, accepted & 0.025945 & 0.065726 & +1.21 \\
Shared MAE, accepted & 0.025420 & 0.064348 & +3.21 \\
Leadwise MSE, accepted & 0.026262 & 0.066494 & +0.00 \\
Leadwise MAE, accepted & 0.025353 & 0.064398 & +3.46 \\
\bottomrule\end{tabular}
\end{table}

\subsection{Does MAE Alignment Help the Accuracy Target?}
Under the common raw first-horizon tree constraint, MSE-to-MAE fitting
reduces sMAE by 0.000440 for shared weights and 0.000762 for
horizon-specific weights (Table~\ref{tab:contrasts}). Both pointwise
14-day intervals exclude zero. Figure~\ref{fig:factorial}a also reports
the four-contrast Bonferroni sensitivity intervals: the shared contrast
then spans zero, whereas the horizon-specific contrast remains positive.
The favorable aggregate point estimates should not be read as uniform
evidence across parameterizations and uncertainty conventions.

All horizon-specific-MSE policies are vetoed to the tree. Only one
shared-MSE seed is accepted, and that seed slightly outperforms its
shared-MAE counterpart. Selection and confirmation therefore do not make
MAE fitting superior in every run. Complete full, raw and accepted scores
are reported together in Supplementary Section 1.
\begin{table}[htbp]
\centering\small
\caption{Paired sMAE reductions. Circular UTC-day block resampling uses 14-day blocks and 10,000 draws. Intervals are conditional, exploratory and unadjusted for multiple comparisons or research selection.}\label{tab:contrasts}
\begin{tabular}{lrr}\toprule
Contrast & Reduction & 95\% interval \\\midrule
MAE gain, shared & 0.000440 & [0.000062, 0.000752] \\
MAE gain, leadwise & 0.000762 & [0.000459, 0.001030] \\
Leadwise gain, MSE & -0.000242 & [-0.000316, -0.000182] \\
Leadwise gain, MAE & 0.000080 & [-0.000012, 0.000182] \\
Leadwise gain, MAE (accepted) & 0.000066 & [0.000010, 0.000124] \\
\bottomrule\end{tabular}
\end{table}

\subsection{Are More Weight Parameters Needed?}
With MSE fitting, horizon-specific weights worsen raw sMAE by 0.000242
relative to shared weights. With MAE fitting, they improve it by only
0.000080, with a 14-day interval spanning zero. The 7-day interval is
slightly positive and the 28-day interval spans zero. Evidence for the
additional 105 free coefficients is therefore sensitive to temporal
resampling, and no practical-equivalence margin was specified in advance.

\begin{figure}[htbp]
\centering
\includegraphics[width=\linewidth]{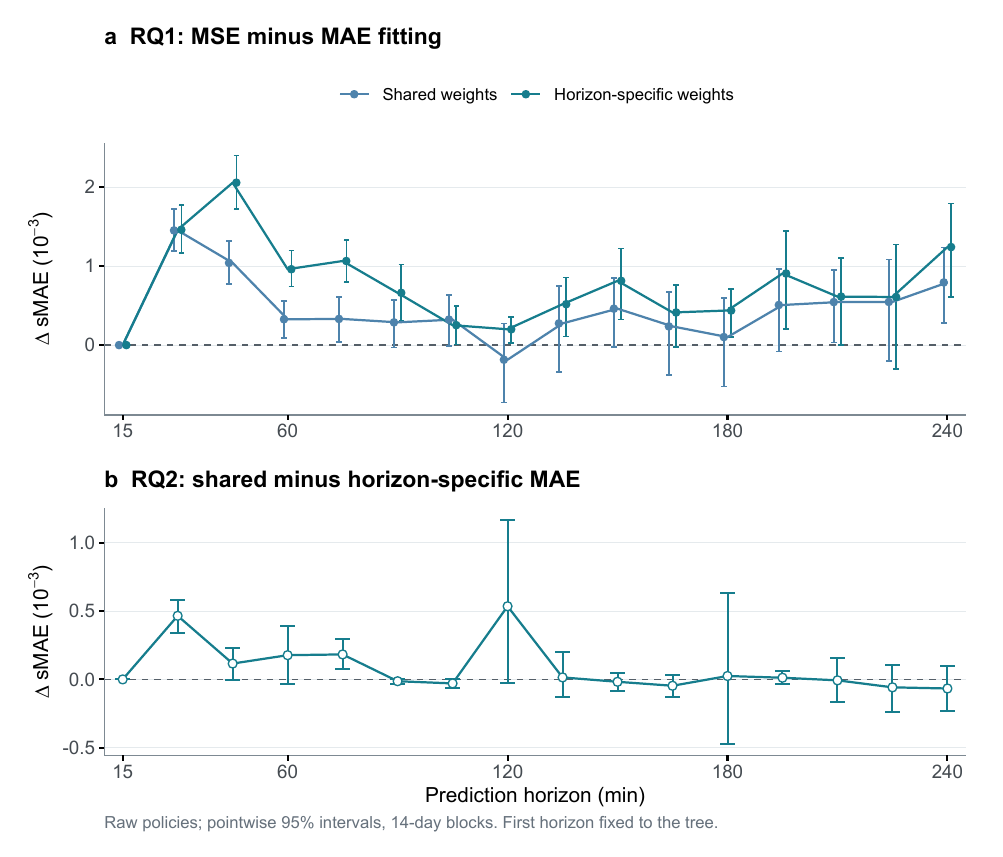}
\caption{Paired sMAE reductions at all 16 horizons. Positive values favor
MAE fitting in (a) and horizon-specific MAE weights in (b). Raw policies
share the first-horizon tree constraint, so the first zero is imposed,
not estimated. Points average seed-specific losses. Bars are conditional
pointwise 95\% intervals from 14-day UTC blocks and are not adjusted for
screening horizons. The curves are exploratory and do not define a
switching threshold.}
\label{fig:horizons}
\end{figure}

Figure~\ref{fig:horizons} separates the loss-alignment and weight-granularity
contrasts across all 16 horizons. The latter is not uniformly favorable
to horizon-specific weights. No horizon is selected from these curves
for the main claim. After policy selection and confirmation, the accepted
horizon-specific MAE score is 0.000066 lower than the shared score. That
comparison includes policy decisions and does not isolate coefficient
dimensionality.

\subsection{Member Error and Realized Cancellation}
\begin{figure}[htbp]
\centering
\includegraphics[width=\linewidth]{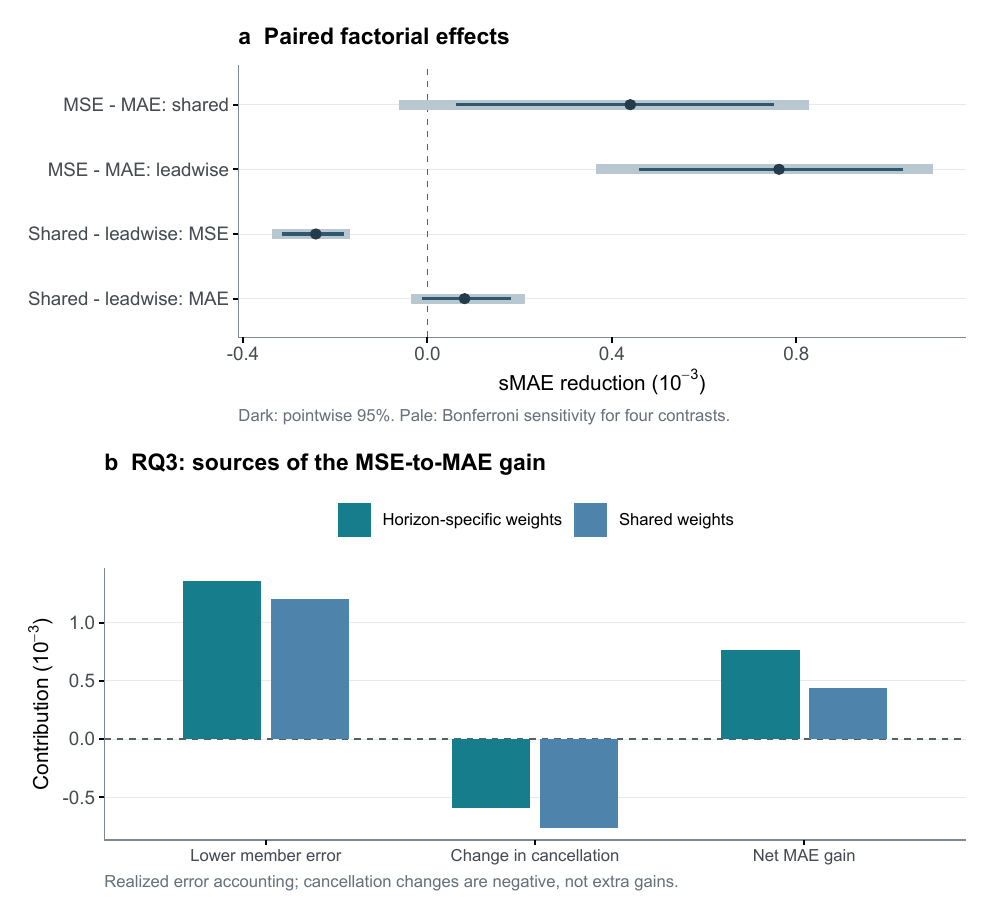}
\caption{Factorial evidence and error accounting for raw policies.
(a) Overall paired effects using 14-day blocks. Dark intervals are pointwise 95\% intervals; pale intervals show Bonferroni sensitivity for four contrasts. This retrospective adjustment does not account for earlier research decisions. (b) The MSE-to-MAE gain equals $(A_{\rm MSE}-A_{\rm MAE})+(C_{\rm MAE}-C_{\rm MSE})$. Cancellation contributes negatively under both parameterizations. Bar heights are descriptive point estimates with no uncertainty bars.}
\label{fig:factorial}
\end{figure}

For shared weights, moving from MSE to MAE fitting reduces weighted member
error $A$ by 0.001200 but also reduces cancellation $C$ by 0.000761,
yielding a net sMAE reduction of 0.000440. For horizon-specific weights,
the corresponding reductions are 0.001354 and 0.000591, yielding
0.000762 (Table~\ref{tab:accounting}; Fig.~\ref{fig:factorial}b).
The reduction in weighted member error exceeds the loss of cancellation, producing the aggregate gain. These quantities describe realized loss, not changes in the
individual frozen predictors or evidence of a causal mechanism.
\begin{table}[htbp]
\centering\small
\caption{Descriptive A/C accounting relative to the tree for the raw factorial policies. A negative difference A minus C denotes improvement. More cancellation alone is not the objective.}\label{tab:accounting}
\begin{tabular}{lrrr}\toprule
Family & $A$ & $C$ & $A-C$ \\\midrule
Shared MSE & 0.002023 & 0.002422 & -0.000399 \\
Shared MAE & 0.000823 & 0.001661 & -0.000839 \\
Leadwise MSE & 0.002256 & 0.002413 & -0.000157 \\
Leadwise MAE & 0.000903 & 0.001822 & -0.000919 \\
\bottomrule\end{tabular}
\end{table}

\subsection{Robustness and Sensitivity}
Within-sample system and seasonal summaries are reported in Supplementary
Section 4. The shared-weight MSE-to-MAE gain
reverses in summer (sMAE reduction $-0.000205$), despite a positive annual
average. Horizon-specific MAE improves over shared MAE in system 34
($0.000146$) but slightly worsens it in system 35 ($-0.000010$), and
also worsens it in winter. These descriptive subgroups are neither
independent replications nor evidence for a subgroup-specific policy.

The optional-prefix control permits zero tree-only horizons. One seed
then selects that option; mean horizon-specific-MAE sMAE is 0.025356,
close to 0.025353 with a compulsory prefix. This weakens a universal
tree-first rule. It does not identify when a recursively applied tree
would fail: every member here makes direct multi-horizon predictions.
Supplementary Sections 3 and 4 report the prefix choices and sensitivity of the intervals to 7/14/28-day blocks, including unfavorable contrasts.

\subsection{Matched Public Retraining without Future Weather}
The latest retraining check examines the same question with an operationally simpler input set. It retains the PVDAQ systems and frozen chronological splits, making it a protocol check on an exposed period rather than a new external validation. All neural parameters are initialized anew; no private pretrained weights or station embeddings are imported. Inputs comprise the 96 power observations up to the origin, known future calendar/solar geometry, and available target-aligned power lags from one, two and seven days earlier. Weather values and their validity masks are zero. The direct tree receives the same raw power history, lag information and known geometry as the full pipeline. Power at the origin is assumed to be available at issue time; the experiment does not establish actual telemetry latency.

The base experts use 20,053 training windows, fusion uses 4,311 windows, and optional joint adaptation uses 3,411 windows.
The standalone comparison tree is refitted on their union of 27,775
windows, so it also receives the adaptation labels. Epochval selects base checkpoints and tree boosting rounds; Confirmation selects the optional correction strength from $\{0,0.25,0.5,1\}$. Test targets do
not enter any of these decisions. For seeds 201, 202 and 203, each neural
candidate is trained for eight epochs of 64 station-balanced batches
of 128 examples, with learning rate $3\times10^{-4}$. The tree uses an L1 objective, 31 leaves, a learning rate of 0.05, at most 500 boosting rounds and 40-round early stopping. The comparison matches information, label access and seed count, but not parameter count or computation. This fixed
budget does not establish convergence of every candidate.

This check averages the three seed predictions \emph{before} scoring.
Its primary metric is daylight MAE averaged equally over the two systems:
\begin{equation}
 L_{\rm day}=\frac12\sum_{s\in\{34,35\}}
 \frac{\sum_{(i,h)\in D_s}|\overline q_{ih}-y_{ih}|}{|D_s|},
 \qquad \overline q=\frac13\sum_{r=1}^{3}q^{(r)},
\end{equation}
where $D_s$ contains geometrically daylight targets for system $s$.
The numerical station scales remain 134.0 and 116.6~kW. These are fixed scaling constants inherited from the prepared data, not independently verified nameplate capacities. All methods use the same daylight mask, geometry-night zeroing and $[0,1.25]$ clipping. The aggregation and feature choices differ from those in the historical experiment; its 3.21\% reduction and the reduction in this check must therefore remain separate results.

\begin{table}[htbp]
\centering\small
\caption{Matched public retraining check for the exposed 2016 period.
Daylight MAE is computed with equal-system weighting after averaging predictions.
The independent tree receives all training labels available to the full
pipeline. Percent reductions refer only to this table's primary metric.}
\label{tab:matched-public}
\begin{tabular}{@{}p{0.38\linewidth}rrr@{}}
\toprule
Method & $L_{\rm day}$ & \shortstack{Daylight MAE\\(kW)$^*$} & \shortstack{Reduction\\(\%)} \\
\midrule
Matched-label standalone tree & 0.05086079 & 6.478810 & 0.00 \\
Static tree-anchored combination & 0.04995439 & 6.372286 & 1.78 \\
Combination + bounded joint correction & 0.04995897 & 6.372638 & 1.77 \\
\bottomrule
\end{tabular}
\par\smallskip\footnotesize $^*$The kW column pools daylight targets
across systems and is a separate aggregation, not $L_{\rm day}$ multiplied
by a common capacity.
\end{table}

Table~\ref{tab:matched-public} supports the static combination as the
simpler MAE-focused design. Later-horizon tree weights are 0.5849,
0.5292 and 0.5554 for the three seeds. Only PatchTST, N-HiTS-PV and
TimeXer-PV receive positive neural weights in these fits; TiDE, LSTM,
the TCN and TimeMixer-PV receive zero static weight. This does not show
that the inactive architectures are generally ineffective. It shows
that this dataset and fitting recipe do not justify an equally weighted
or obligatorily eight-active-branch interpretation.

The latest optional extension jointly updates TimeMixer-PV and PatchTST
and computes a bounded residual correction. It leaves the first horizon,
geometrical night and baseline forecasts below 0.01 unchanged, and caps
the correction at 0.005 in scaled units. Its four epochs of 64 batches
use a combined absolute/squared-error objective with a risk penalty,
not pure MAE. Confirmation selects strengths 0.5, 0 and 0. Aggregate MAE is slightly higher than for the static combination, although the RMSE-derived score improves slightly. This extension is therefore a negative incremental result, not the source of the paper's MAE gain.

\section{Discussion and Limitations}
\subsection{What the Accuracy Result Supports}
The evidence supports a strong tree combined with selectively weighted temporal predictions. In the historical bank, single-expert selection retains the tree at every horizon, whereas equal averaging performs worse. An MAE-aligned combination can nevertheless improve on the tree, indicating that the best standalone forecast does not exhaust the useful information in the bank. Shared weights recover most of the historical accepted gain without the 105 extra free coefficients required by horizon-specific fitting. Matched-label retraining gives the same qualitative result without future weather. These findings justify testing a combiner; they do not establish that eight experts are optimal or that every member is necessary.

The tree anchor, MAE objective and simplex weights that permit zeros address the same accuracy problem. A strong tabular baseline limits weak comparisons, fitting to MAE targets the reported error, and zero weights limit dilution by weak candidates. Protecting the first lead remains a conservative implementation choice, but the optional-prefix results do not support a universal tree-only rule for a fixed initial duration. Additional routing, attention or correction modules must demonstrate their own incremental MAE evidence. The latest joint correction fails that test in the present retraining check.

\subsection{What Remains Unresolved}
We have not established that cloud fluctuations, ramp events or a
particular meteorological variable cause the remaining errors. The
historical data use retrospective ERA5, and the retraining check removes
weather inputs entirely. A claim of weather-specific robustness would
require an issue-time weather archive, prespecified regime definitions
and error comparisons within those regimes. Adding a weather module
solely because it is physically plausible would not answer the present
MAE question.

Both systems are co-located and both analyses reuse an exposed evaluation
period. The historical tree is shared across neural seeds; the retraining
tree also has effectively identical fits under the fixed deterministic
recipe. Overlapping targets and seeds are not independent site replications.
Historical conditional intervals do not include the sequence of research
decisions; subgroup reversals and adjustment sensitivity limit claims of
stable improvement. Different aggregation, inputs and fitting budgets
prevent interpreting the two tables as a numerical replication or a
general neural-versus-tree ranking. The matched retraining check supplies
no future-weather advantage, but does not validate real telemetry latency
or deployment performance.

The next confirmation should freeze the static tree-anchored MAE
combination and its matched-label tree comparator before examining a new
site or time period. It should test whether the residual MAE reduction
exceeds a prespecified practical threshold. Member-removal tests are
needed before claiming an indispensable expert, and published-time
weather evidence is needed before claiming weather-driven improvement.

\section{Conclusion}
This study addresses one problem: reducing the residual MAE of a strong
direct tree over 15--240-minute photovoltaic power forecasts. A
tree-anchored, shared-MAE combination lowers historical scaled MAE by
3.21\%. A separate public retraining check without future weather lowers
daylight, equal-system MAE by 1.78\% against a tree with the same
training-label access. The protocols differ, and both reuse an exposed period. Within this bank, the results support selective combination over indiscriminate averaging. They do not establish a need for eight active experts, horizon-specific weights or the latest joint correction. Independent confirmation is needed before claiming transferable accuracy gains or an operational weather-forecasting benefit.

\section*{Data and Code Availability}
PVDAQ is publicly available \citep{deline2021pvdaq}. Frozen predictions,
configuration records and non-fitting audit scripts are retained locally.
The matched retraining records include a frozen feature/split protocol,
code and data hashes, three seed predictions, an independent metric audit
and hashes for 126 remote model/configuration artifacts. The machine-readable
retraining protocol calls the fixed station scales capacities; the
manuscript conservatively uses the inherited prepared-data scaling
definition stated above. Public archival release and redistribution
checks remain pending; no public reproducibility deposit is claimed.

\bibliographystyle{elsarticle-harv}

\end{document}